\documentclass[runningheads]{llncs}
\usepackage[utf8]{inputenc}
\usepackage[small]{caption}
\usepackage{graphicx}
\usepackage{amsmath}
\usepackage{booktabs}
\usepackage{url}

\usepackage{breakurl}
\usepackage[breaklinks]{hyperref}
\usepackage{adjustbox}
\usepackage{tikz}
\usetikzlibrary{tikzmark,calc}
\usetikzlibrary{arrows,automata,positioning,decorations.markings}
\usetikzlibrary{fit,shapes,shapes.multipart, matrix,arrows.meta}
\usepackage{booktabs}
\usepackage{amsfonts}
\usepackage{nicefrac}
\usepackage{microtype}
\usepackage{amssymb}
\usepackage{bbm}
\usepackage{multirow}
\usepackage{threeparttable}
\usepackage[noend]{algpseudocode}
\usepackage{algorithmicx,mathtools,algorithm,eqparbox,array}

\begin{document}
\algnewcommand{\LineComment}[1]{\State // #1}
\title{Neural-Symbolic Relational Reasoning on Graph Models: Effective Link Inference and Computation from Knowledge Bases\thanks{Supported by Coordenação de Aperfeiçoamento de Pessoal de Nível Superior - Brazil (CAPES) - Finance Code 001 and by CNPq - the Brazilian Research Coucil}}
\titlerunning{Neural-Symbolic Relational Reasoning on Graph Models}
%
\author{Henrique Lemos\inst{1}\orcidID{0000-0003-0236-1291} \and
Pedro Avelar\inst{1}\orcidID{0000-0002-0347-7002} \and
Marcelo Prates\inst{1}\orcidID{0000-0002-5576-7060} \and
Luís Lamb\inst{1}\orcidID{0000-0003-1571-165X} \and
Artur Garcez\inst{2}\orcidID{0000-0001-7375-9518}
}
\authorrunning{H. Lemos et al.}
%
\institute{Institute of Informatics, UFRGS, Porto Alegre - Brazil
\email{\{hlsantos,morprates,phcavelar,lamb\}@inf.ufrgs.br}
\and
Department of Computer Science,
City, University of London, London - UK
\email{a.garcez@city.ac.uk}
}
\maketitle              
\begin{abstract}
The recent developments and growing interest in neural-symbolic models has shown that hybrid approaches can offer richer models for Artificial Intelligence. 
The integration of effective relational learning and reasoning methods is one of the key challenges in this direction, as neural learning and symbolic reasoning offer complementary 
characteristics that can benefit the development of AI systems. 
Relational labelling or link prediction 
on knowledge graphs has become one of the main problems in deep learning-based natural language processing research. Moreover, other fields which make use of neural-symbolic techniques may also benefit from such research endeavours. 
There have been several efforts towards the identification of missing facts from existing ones in knowledge graphs. Two lines of research try and predict knowledge relations between two entities by considering all known facts connecting them or several paths of facts connecting them. 
We propose a neural-symbolic graph neural network which applies learning over all the paths by feeding the model with the embedding of the minimal subset of the knowledge graph containing such paths. 
By learning to produce representations for entities and facts corresponding to word embeddings, we show how the model can be trained end-to-end to decode these representations and infer relations between entities in a multitask approach. Our contribution is two-fold: a neural-symbolic methodology leverages the resolution of relational inference in large graphs, and we also demonstrate that such neural-symbolic model is shown more effective than path-based approaches.

\end{abstract}
\section{Introduction}
\label{sec:int}
Neural-symbolic computing has recently become one of the promising research subfields in artificial intelligence, with both academic researchers and companies such as IBM and Microsoft setting up agendas that foster the integrated use of connectionist learning and symbolic inference techniques \cite{garcez19,raghavan19,smolensky19,mao19}. 
Deep learning (DL) models are now well-established frameworks towards solving Natural Language Processing (NLP) tasks. In recent years, however, the need for improved explainability and interpretability in machine learning and AI, has called for the investigation of models which lend themselves to clear semantic analyses, which in turn suggest the use of neural-symbolic approaches \cite{evans18,garcez19,mao19}. 

In this work, we propose a neural-symbolic graph neural network (GNN) model to perform integrated relational learning and inference over large scale knowledge graphs. To show the effectiveness and generality of our approach, we solve the relevant task of reasoning and inference on link prediction. 
More specifically, we leverage the relational learning capabilities of graph neural networks, which lend themselves soundly to integrated learning and reasoning tasks, which one can claim is the \emph{raison d'etre} of neural-symbolic models. The core concept of our neural-symbolic model is built upon a connectionist architecture modelled by multilayer perceptrons (MLP) and long-short term memories (LSTM) whose message-passing structure reflects the relational knowledge expressed by the graph itself.

In this context, knowledge bases (KBs) are understood as  a repository of \emph{facts} stated over pairs of \emph{entities}. Facts can be described by 3-tuples $(e_s,r,e_t)$ connecting a source entity $e_s$ and a target entity $e_t$ with a relation $r$. For example, the fact that \emph{the LHR Airport serves the city of London} can be formalised as (LHR Airport, serves, London). Because KBs are often compiled from scraped data, they may be incomplete, with missing facts about the stored entities. This motivates the problem of link prediction, in which one learns to predict previously unknown facts between two entities given the existing ones in the KB. Traditionally, attempts to solve this problem have relied on accumulating information over all facts directly connecting two entities to predict a new one \cite{bordes2013,socher2013,nguyen2016}. This approach is conceptually limited, as two entities can possibly be connected by \emph{paths} of facts and these paths may store useful information which could also be exploited to predict new facts. The most obvious benefit from considering this kind of input is also that eventually there will be no direct relation between two entities in the KB and yet we may infer some relation between them \cite{lao2011,neelakantan2015,yin2018}. A recent study has made an effort to take these paths into consideration, by processing each path connecting a source and a target entity via a RNN and accumulating their outputs to produce a prediction \cite{das2017}. 

Our approach goes a step further by acknowledging that useful information can be lost when one does not take all paths into consideration \emph{at the same time}. To overcome this issue, we propose a neural-symbolic graph neural network model which is able to feed on the minimal sub-graph containing all paths connecting a given source and target pair (fully described in Section \ref{sec:model}). A GNN is an end-to-end differentiable DL model which feeds on graphs. It does so by assigning multidimensional embeddings\footnote{We use the words ``embeddings'' and ``representations'' depending on the context, but with the same meaning of real-valued projections of some object.} for each node and each edge in the graph and refining these embeddings over many iterations of a parameterised (and thus trainable) procedure of ``message-passing''. Techniques in the GNN family have enjoyed increased interest in the last years, with applications to generative models of source code \cite{brockschmidt2018generative}, quantum chemistry \cite{gilmer2017neural} and urban planning \cite{hu2018recurrent}.

The remainder of the paper is structured as follows. Next, we formalise the problem tackled in the paper and briefly discuss related work. In  Section~\ref{sec:model}, describe our neural-symbolic model and how we trained it on link prediction. Finally, in Section~\ref{sec:res} and Section~\ref{sec:conc}, we analyse and compare our results to state-of-art techniques and point out directions for further research.

\section{Preliminaries}
\label{sec:prob}
The problem of link prediction can be formalised as follows. A Knowledge Base (KB) is a 4-tuple ${\mathbf{KB}} = (\mathcal{E},\mathcal{T},\mathcal{F},\mathcal{R})$, where $\mathcal{E}$ is the set of entities, $\mathcal{T}$ is the set of entity types, $\mathcal{F}$ is the set of facts and $\mathcal{R}$ is the set of relations. Given a subset ${\mathbf{KB}' = (\mathcal{E}',\mathcal{T}',\mathcal{F}',\mathcal{R}') \subseteq \mathbf{KB}}$ of a knowledge base ${\mathbf{KB} = (\mathcal{E},\mathcal{T},\mathcal{F},\mathcal{R})}$ and two entities $e_s, e_t \in \mathcal{E}'$, we want to predict a fact $f=(e_s,r,e_t)$ such that $f \not\in \mathcal{F}'$ but $f \in \mathcal{F}$, that is, we want to use the (incomplete) information in $\mathbf{KB}'$ to recover facts from its superset $\mathbf{KB}$.

It should be noted that, in general, $\mathbf{KB}$ is not known. The motivation for link prediction is precisely the fact that knowledge bases are usually incomplete, so that our goal is to learn a procedure for enriching an input knowledge base into its (hypothetical) most complete form. In practice, for a given real-world knowledge base, there is no way for one to evaluate whether the output of a link predictor is correct, because the missing labels are not known. But one can still train such a model in a supervised way by training it on subsets of real-world knowledge graphs and forcing its outputs to match the removed data.

On the subject of link prediction, the \textit{Universal Schema}~\cite{riedel2013} can be seen as the seminal approach, tackling the problem by enforcing the learning of latent feature embeddings for entity tuples and relations through matrix factorisation. This and other initial efforts rely solely on the set of facts directly connecting two entities in order to predict a new one \cite{bordes2013,socher2013,nguyen2016}. This can be improved by taking \emph{paths} of facts into consideration: apart from being directly connected by a set of facts, two entities can also be indirectly connected by a sequence of them (for example ``the \textbf{LHR Airport} is located in \textbf{England} of which the capital is \textbf{London}''). Considering this approach, one may highlight the Path Ranking Algorithm (PRA)~\cite{lao2011} and the Path-RNN~\cite{neelakantan2015}. One may also highlight the work of Das et al.~\cite{das2017} which enhanced the task of reasoning over chains of relations, previously performed by PRA and Path-RNN, by including representations also for the entities and for the entity types along the path, and by considering multiple paths, combined via score pooling methods, to predict the missing relation between source and target entity. More recently, Yin et al.~\cite{yin2018} tackled the same issue by encoding the paths with Gated Recurrent Units and by updating the embeddings multiple times along the entire path. These four methods were jointly evaluated on the dataset released by Das et al.~\cite{yin2018} and the latter has achieved the best results so far. As we will see shortly, our approach extends the concept of learning over \emph{paths} to learning over \emph{graphs}. In this context, the evolution of approaches toward link prediction can be summarised as 1) learning over \emph{edges}, 2) learning over \emph{paths} and 3) learning over \emph{graphs} (this paper).

The \emph{Single-Model} introduced by Das et al.~\cite{das2017} is parameterised by three trainable components: the first is a table of representations (i.e. real-valued vectors) for each entity in the KB, the second is a table of representations for each relation in the KB, and the third is a Recurrent Neural Network (RNN) which will be fed with a sequence of facts $\pi = e_s,r_0,e_1,r_1,e_2,\dots,r_n,e_t$ connecting a source entity $e_s$ with a target entity $e_t$. The RNN's hidden state can be seen as a representation for the path $\pi$, which can be later decoded into a prediction for a new fact connecting $e_s$ and $e_t$ directly. At each timestep, the input for the RNN is a tuple $(e_i,r_i)$ where $e_i$ is the i-th entity in the path and $r_i$ is the relation connecting it with the following entity. Because there is no relation later to $e_t$, the last input $(e_t,r_{dummy})$ contains a ``dummy'' relation:

\begin{equation}
\begin{aligned}
    \mathbf{h}^{(0)} &\leftarrow f(\mathbf{h}^{(s)},(e_s,r_0)) \\
    \mathbf{h}^{(i+1)} &\leftarrow f(\mathbf{h}^{(i)},(e_i,r_i)) \\
    \mathbf{h}^{(last)} &\leftarrow f(\mathbf{h}^{(n)},(e_t,r_{dummy}))
\end{aligned}
\end{equation}

For each relation $r$, a similarity measure is computed between the representation for the path $\pi$, produced by the RNN, and the representation for $r$, which is learned. But because the RNN can only process each path separately, the authors propose finding different paths using random walks in the KB and score-pooling the outputs to accumulate their information.

\begin{figure}[t!]
    \centering
    \begin{adjustbox}{width=12.5cm}
    \includegraphics[width=0.9\linewidth]{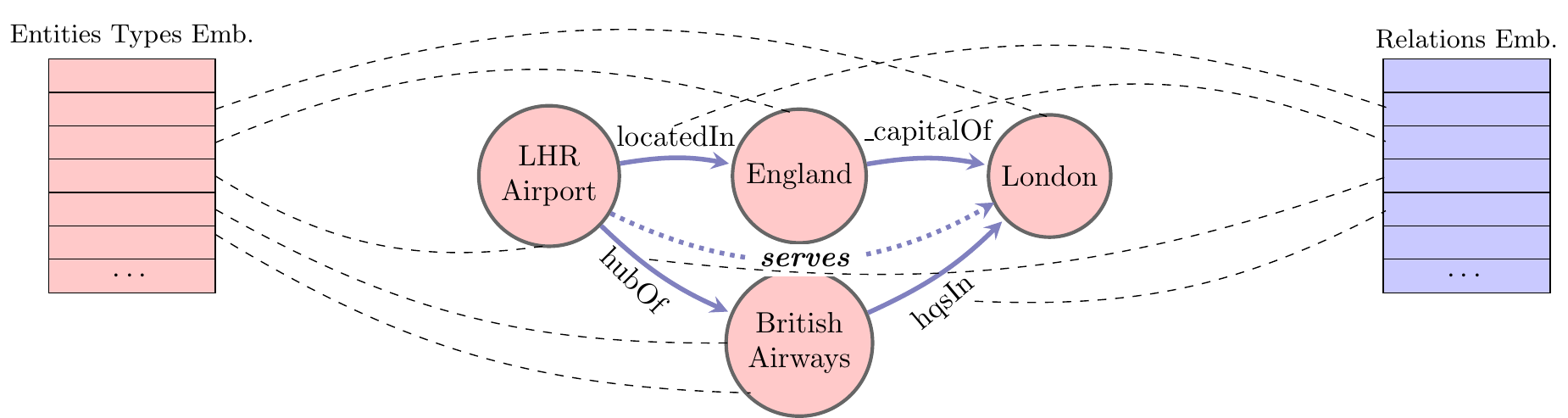}
     \end{adjustbox}
    \caption{A dummy example of how a knowledge graph between two entities (LHR Airport and London) is mapped into two lookup tables. Each relation is mapped into a unique relation embedding, while each entity may be mapped into several entity types (e.g. British Airways node), which are then fed to an aggregation function, such as mean or sum. Note that {$\_$}capitalOf is an inverse relation, so it is read backwards.}
    \label{fig:initialization}
\end{figure}

The main enhancements of the \emph{Single-Model}, compared to PRA and Path-RNN, are the addition of embeddings to represent the entities along the path, the inclusion of neural attention mechanisms (in terms of score-pooling) to reason over multiple paths and the ability to train one single model to predict several target relations. However, later work of Yin et al.~\cite{yin2018} was able to improve the performance on the dataset provided by Das et al.~\cite{das2017}. Their strategy consisted in forcing the RNN (a Gated Recurrent Unit - GRU) to predict entities representations as outputs along the path, and to update these representations at each step of the path. One may also highlight the work of Xiong et al.~\cite{xiong2017} which defines a policy-based agent to reason over a KG according to a reward function -- at each step the agent picks a relation to extend the KG it is interacting with.
These two approaches understand the problem of link prediction as combining multiple paths connecting $e_s$ to $e_t$, and learning their intermediate representations separately. By changing the perspective to a graph-view, one could expect a faster learning since relations and entities of different paths will be aware of each other when considered inside the same graph. 

Only very recently have link prediction tasks been tackled by neural networks whose inputs are graphs. Outside of a KB scope, work of Zhang and Chen~\cite{zhang2018} have already demonstrated unprecedented performance of GNNs on the binary link prediction task -- with no edge features. Still, the R-GCN model~\cite{schli2018} relies on a graph convolution architecture to perform link prediction on KGs: the R-GCN itself can be seen as an encoder, which takes the graph as an input and outputs a real-valued vector for each vertex $v_i \in \mathcal{V}$; then a decoding step (\textit{DistMult} algorithm) scores 3-tuples through a function whose inputs are the real-valued vectors of the source entity and the target entity, and a learned diagonal matrix associated with the relation. The crucial difference of our model to the R-GCN is the statefulness:
while their models learn functional applications over a neighbourhood, treating the node in the same way it treats its neighbours, our model uses LSTMs to process the abstractions learned in a way that the node uses its own state and its neighbourhood aggregation.

\section{A Neural-Symbolic Model for Reasoning on Graphs}
\label{sec:model}

As stated before, our model goes one step further than previous approaches and trains a model to reason directly over \emph{graphs} as opposed to \emph{paths}. In this context, before we fully describe our model, it is important to clarify how the information in the KB is modelled as a graph. First of all, our model is not fed with the \emph{entire} KB graph but with subsets thereof. Such a subset corresponds to the minimal subgraph containing all known paths connecting the source and target entities of interest $(e_s,e_t)$. Entities correspond to \emph{nodes} in our graph, while facts correspond to directed \emph{edges}. Each edge is \emph{labelled} with its corresponding relation, and each node is labelled with the set of entity types it belongs to. Also, because any two entities can be directly linked by more than one fact, we have \emph{parallel edges}. In summary, our learning task is defined over a (node and edge)-labelled directed graph with parallel edges.

Over the course of its computation, our model will refine (real-valued vector) representations for each entity and for each fact in our graph. Because entities (nodes) and facts (edges) are labelled, the initial representation for each of them must be initialised with the corresponding label -- note that we also force a \emph{fake} fact (edge) between $e_s$ and $e_t$ whose embedding is initialised with zeros. In this context, the first two trainable components of our model are two embedding layers, which can be conceptualised as two tables, the first one storing representations for entity types and the second one storing representations for relations. These representations will be used to initialise the representations for entities and facts. The reader should be reminded that each entity can belong to possibly many entity types at once. So, in our model, the initial representation for each entity is computed as an aggregation of the representations of all entity types it belongs to. As each fact is associated with a single relation, fact representations are simply initialised with the representation for their corresponding relations (Figure \ref{fig:initialization}). This process is described throughout lines 8--9 of Algorithm \ref{alg:GNN}.

\begin{figure}[t!]
    \centering
    \begin{adjustbox}{width=12cm}
     \includegraphics[width=0.9\linewidth]{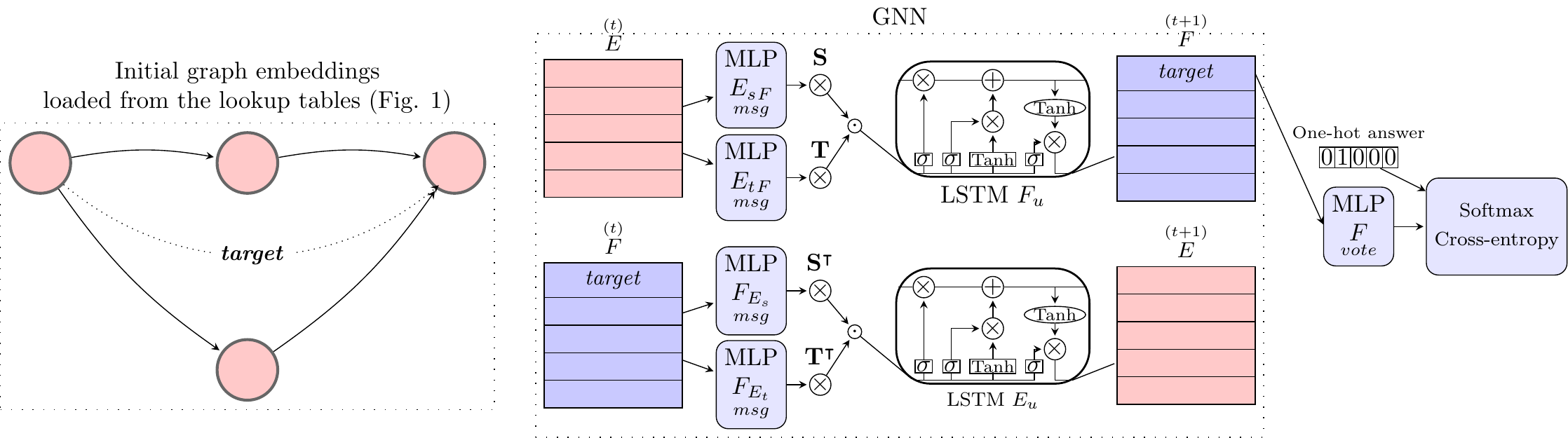}

     \end{adjustbox}
    \caption{Overall view of our model: the initial embeddings are loaded from the lookup tables -- except the \textit{target} relation embedding  which is filled with zeros -- and then fed to the GNN internal memory ($E^{(t)}$ and $F^{(t)}$). After \textit{t} timesteps, we gather the updated \textit{target} relation embedding and provide it as the input to a voting MLP, which will answer which relation is being represented by this embedding. The loss computed by Softmax Cross-entropy is then backpropagated throughout all the neural modules (MLP, GNN and embedding layer). Note that we omit the hidden and cell states and the unrolling of the two LSTMs inside the GNN to optimise readability.}
    \label{fig:model}
\end{figure}

Then, the kernel of our model is a Graph Neural Network. As briefly mentioned above, a GNN can be seen as an end-to-end differentiable, message-passing algorithm between graph elements which is implemented with neural components. A GNN initially assigns (real-valued vector) representations for nodes and edges, and then iteratively refines these representations through many iterations of message-passing. In each message-passing iteration, the representation for each node is fed to an MLP which computes a representation for a ``message'' which will be sent to its neighbour elements. In the same way, another MLP computes representations for edge messages. These messages will be sent to neighbour elements in the graph: each node receives messages from all edges connected to it, while each edge receives two messages -- one from its source and one from its target node. The set of messages received by each element is aggregated into a single vector -- the matrices multiplications in Figure~\ref{fig:model}) do not only aggregate the messages but also mask them with adjacency information so each node receives an aggregated message only from its neighbours. Finally, for each element, this vector is fed alongside with the current element's representation to an LSTM which updates it into a new representation. This process is described throughout lines 11--17 of Algorithm \ref{alg:GNN}. Also, we avoided to provide a thorough description of how GNNs generally work in this paper, but interested readers may refer to Battaglia et al.~\cite{battaglia2018} and to Wu et al.~\cite{wu2019} for complete and up-to-date surveys of how GNNs work and have evolved along the last years.

The representations for entities and facts are updated by LSTMs. Concretely, we have an LSTM $E_{u}$ to update entity embeddings and an LSTM $F_{u}$ to update fact embeddings. One can see the representations for entities or facts as the \emph{hidden states} of each one of these LSTMs.
Lastly, from line 18 to 20 of Algorithm \ref{alg:GNN}, we gather from the GNN the final embedding of the forced unknown fact and feed it to a \emph{voting} MLP $F_{vote}$. This MLP can also be seen as a decoding function: it receives a fact embedding and extracts its label (relation). It outputs $C$ logits, where $C$ is the total amount of target relations. A \emph{softmax} function is then applied to these logits in order to transform them into $C$ probabilities -- the most likely relation is to be considered as the predicted answer.

\begin{algorithm}[ht]
\caption{Neural-Symbolic Relation Predictor}\label{alg:GNN}
\begin{algorithmic}[1]
\Function{GNN}{\footnotesize ${(\mathcal{E}, \mathcal{T}, \mathcal{F}, \mathcal{R}, \mathbf{M}_{ \mathcal{E\rightarrow T}}, \mathbf{M}_{\mathcal{F\rightarrow R}})}$}
\State 
\LineComment{{\small Compute binary adjacency matrix from facts to source \& target entities}}
\State  $ \mathbf{S}[i,j] \negthickspace \leftarrow \negthickspace \mathbbm{1}  (\exists e', r' | f_i \negthickspace = \negthickspace (e_j,r',e')) |~ \forall f_i \negmedspace \in \negmedspace \mathcal{F}$
\State $ \mathbf{T}[i,j] \negthickspace \leftarrow \negthickspace \mathbbm{1} (\exists e'\exists r' | f_i \negthickspace = \negthickspace (e',r',e_j)) |~ \forall f_i \negmedspace \in \negmedspace \mathcal{F}$

\State
\LineComment{{\small Gather facts \& entity repr. from lookup tables (operator $\langle \rangle$ indicates any aggregation function - i.e. sum, mean, etc.)}}
\State $\overset{(1)}{\mathbf{F}}[i]  \negthickspace \leftarrow  \negthickspace \hphantom{\langle}\mathbf{M}_{\mathcal{F\rightarrow R}}(\mathcal{R}[i]) ~|~ \forall f_i \in \mathcal{F}$ 
\State $\overset{(1)}{\mathbf{E}}[i]  \negthickspace \leftarrow  \negthickspace \langle \mathbf{M}_{\mathcal{E\rightarrow \mathcal{T}}}(\hphantom{\mathcal{R}}\mathllap{\mathcal{T}}[i,k]) \negmedspace ~|~ \negmedspace \forall k \negmedspace \in [0,mt) \rangle ~|~ \forall e_i \in \mathcal{E}$ 

\State
\LineComment{Run $t_{max}$ message-passing iterations}
\For{$t=1 \dots t_{max}$}
  \LineComment{{\small Refine entity repr. with messages from facts}}
  \label{alg:line:vertices_refinement}\State{\footnotesize $\overset{(t+1)}{\mathbf{E}_h}, \negthickspace \overset{(t+1)}{\mathbf{E}} \negthickspace \negthickspace \leftarrow \negthickspace  \underset{u}{E}(\overset{(t)}{\mathbf{E}_h}, \mathbf{S}^\intercal \negthickspace \times \negthickspace \underset{msg}{F_{E_s}}(\overset{(t)}{\mathbf{F}}), \mathbf{T}^\intercal \negthickspace \times \negthickspace \underset{msg}{F_{E_t}}(\overset{(t)}{\mathbf{F}}) ) $ }
  
  \LineComment{{\small Refine fact repr. with messages from entities}}
  \label{alg:line:edges_refinement}\State $ \overset{(t+1)}{\mathbf{F}_h}, \negthickspace \overset{(t+1)}{\mathbf{F}} \negthickspace \negthickspace \leftarrow \negthickspace \underset{u}{F}(\overset{(t)}{\mathbf{F}_h},\mathbf{S} \negthickspace \times \negthickspace \underset{msg}{E_{s_{F}}}(\overset{(t)}{\mathbf{E}}), \mathbf{T} \negthickspace \times \negthickspace \underset{msg}{E_{t_{F}}}(\overset{(t)}{\mathbf{E}}))$
\EndFor

\LineComment{{\small Translate the forced unknown fact (i=0) into logit probabilities of each relation}}
\State $\mathbf{R_{logits}} \leftarrow F_{vote}\left(\overset{(t_{max})}{\mathbf{F}} \small{[0]}\right)$
\LineComment{{\small Apply softmax cross-entropy to identify which is the most likely target relation }}
\State $\textrm{prediction} \leftarrow \textrm{argmax}(\textrm{softmax}( \mathbf{R_{logits}}))$

\EndFunction
\end{algorithmic}
\end{algorithm}

\subsection{Experimental Setup}
Our model architecture is structured as follows. We use embedding sizes of $64$ for entity, fact and message representations. The message-computing MLPs $\mathcal{M}_{\mathcal{E} \rightarrow \mathcal{T}}, \mathcal{M}_{\mathcal{F} \rightarrow \mathcal{R}} : \mathbb{R}^{64} \rightarrow \mathbb{R}^{64}$ are three-layered with layer sizes $(64,64,64)$ and have ReLU nonlinearities as activations for all layers except for a linear activation in the last layer. The update functions $\underset{u}{E}, \underset{u}{F} : \mathbb{R}^{64}, \mathbb{R}^{64} \rightarrow \mathbb{R}^{64}, \mathbb{R}^{64}$ are implemented by layer-norm LSTM cells with ReLU activations. Finally, we define $F_{vote}$ as a four-layered MLP with layer sizes $(64,64,64,46+1)$ with ReLU nonlinearities except for the last linear activated layer.

At the end of the pipeline depicted in Figure~\ref{fig:model}, our model is trained to minimise the following loss~$\mathcal{L}$:

\begin{align}
\label{eq:loss}
\mathcal{L} = & \frac{1}{B} \displaystyle\sum_{b=1}^{B}\left(- \sum_{c=1}^{C} y_{b,c} \log( \frac{ e^{z_{c}} }{ \sum_{c=1}^{C} e^{z_{b,c}} } )\right)
\end{align}

where $B$ stands for the batch size and $C$ for the number of target relations. Note that Equation~\ref{eq:loss} just computes a softmax cross entropy over the $C$ target relations for each instance, averaging over the batch.

Each model was trained for $2000$ epochs, each one comprising $128$ operations of stochastic gradient descent (Tensorflow's Adam optimiser with learning rate $=2 \times 10^{-5}$) in batches of $10$ instances each. We also set $t_{max}$ to $25$ to allow the GNN kernel to refine entities and relations embeddings throughout sufficient iterations

We ran our meta-model under three different configurations:
\begin{itemize}
    \item \textbf{GNN-Relation} -- Only relations were considered, therefore there was only one table to store the embeddings -- $\mathcal{M}_{ \mathcal{F\rightarrow R}}$ -- and all entities were mapped into a single real-valued vector, which is also learned.
    \item \textbf{GNN-Mean} -- Both entities and relations embeddings are mapped into separate tables, and to aggregate several entity types into one entity representation we used arithmetic mean.
    \item \textbf{GNN-Sum} -- The same as \textbf{GNN-Mean} but using sum as the aggregation function over entity types.
\end{itemize}

\ifx
\begin{figure}[ht]
    \centering
    \includegraphics[width=0.6\linewidth]{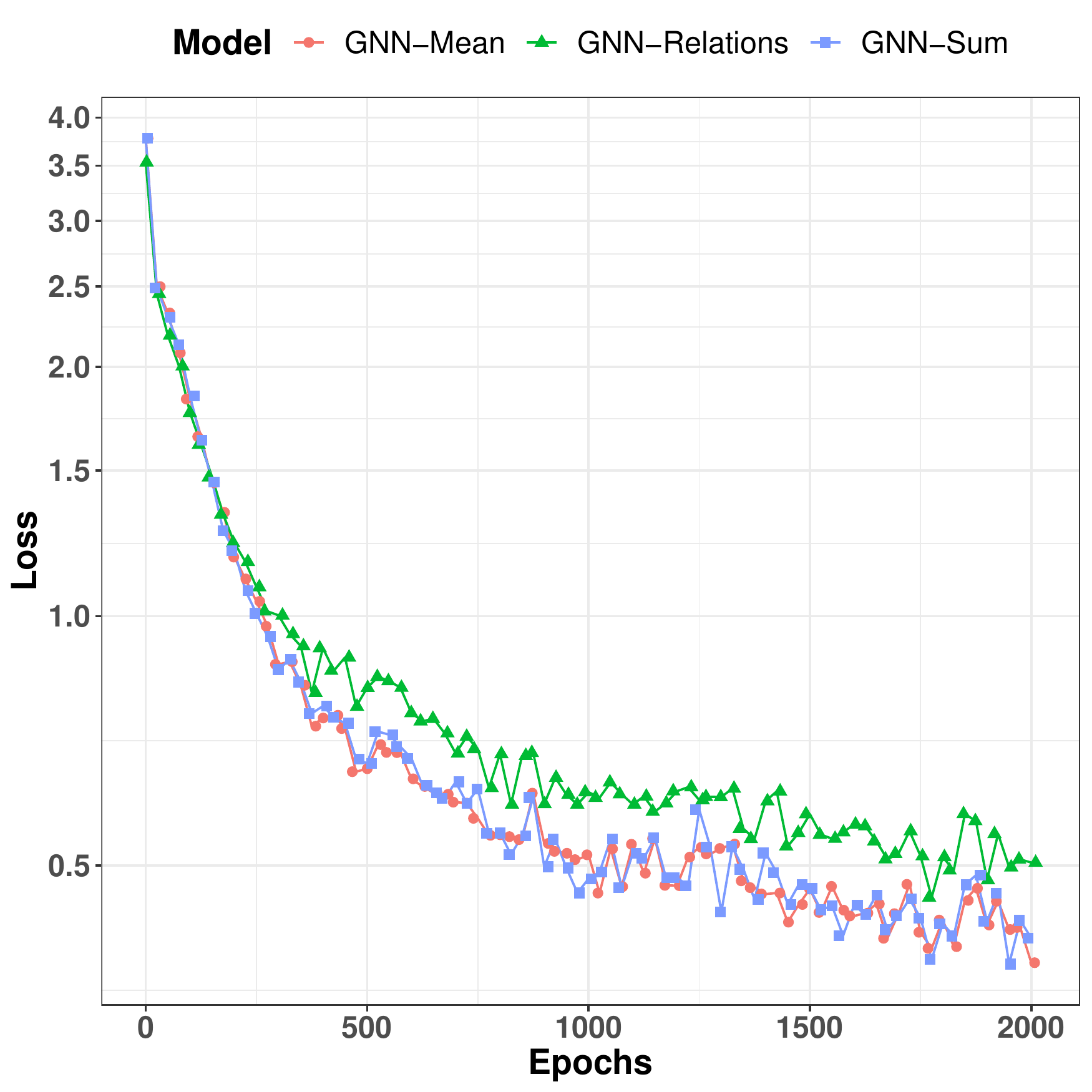}
    \caption{Evolution of the training loss -- Softmax Cross-entropy -- along the training epochs. We observed that incorporating entity types yielded faster learning compared to unlabelled entities.}
    \label{fig:train}
\end{figure}
\fi

During the training procedure, a faster loss decay was observed on \textbf{GNN-Mean} and \textbf{GNN-Sum}, which corroborates the importance of taking both entities and relations into account instead of considering only relations.

\subsection{Benchmark description}
We trained and tested our model on the benchmark dataset released by Das et al.~\cite{das2017}. It comprises $3.22$M entity pairs and $51390$ different types of relations coming either from Freebase or from ClueWeb. We were able to reduce the amount of relation types to $25737$ due to our strategy of incorporating directional data into the graph itself -- a relation and its reversed form, such as "/location/location/contains" and "\_/location/location/contains", were mapped into a single index.

Instead of using an unique representation for each entity, in \textbf{GNN-Mean} and in \textbf{GNN-Sum} we chose to rely  on entities types, as the dataset's distribution of entities is heavy tailed~\cite{das2017}. We gathered this mapping of entity to entity types from the vocabulary provided in the dataset which yielded a total of $2247$ unique entity types. There are positive and negative instances for each one of one the $46$ target relations, all the negative ones were mapped into a \emph{null} relation target, thus resulting in $46+1$ target relations left to be predicted. We used positive and negative instances from both \textit{train} and \textit{dev} files during training, we also forced our minibatch to be balanced by alternating positive and negative (\emph{null} target relation) instances.

\begin{figure*}[t!]
    \centering
    \includegraphics[width=\linewidth,height=6cm,keepaspectratio]{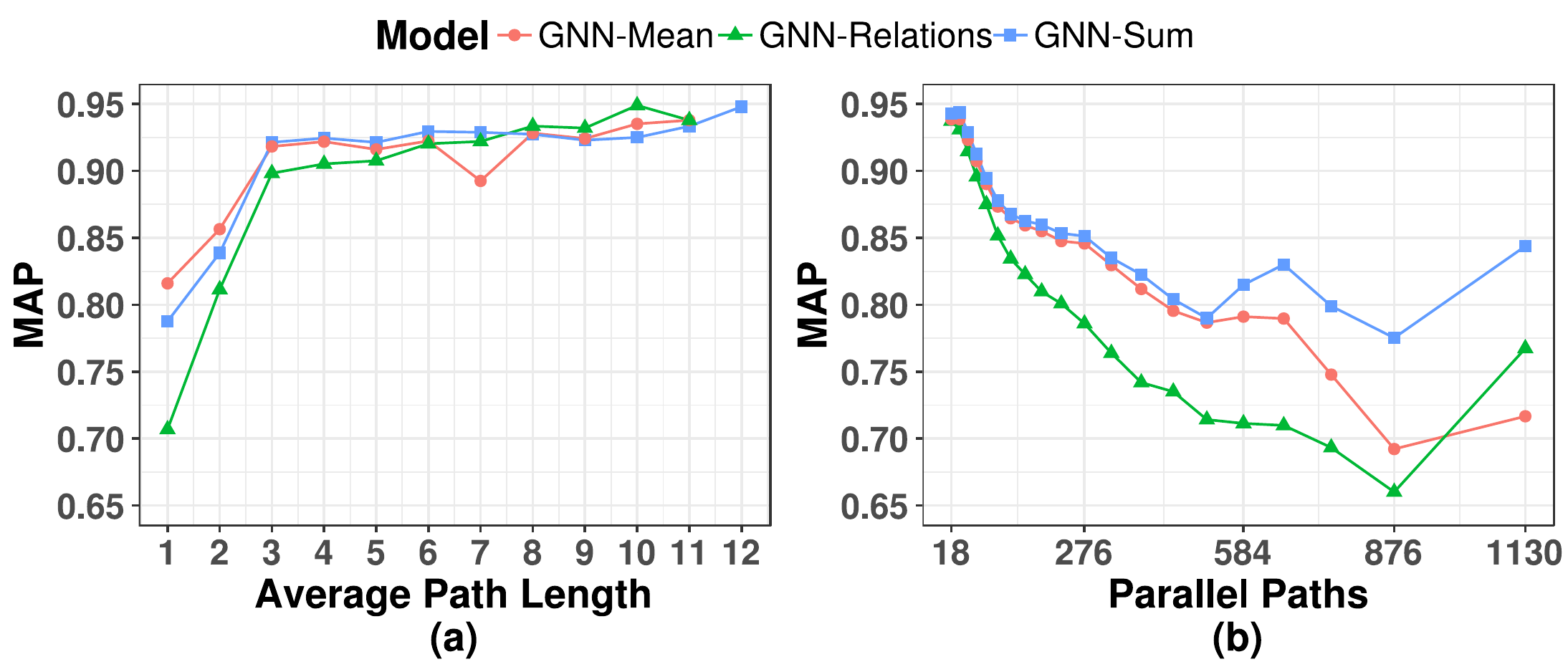}
    \caption{Plot (a): each graph has many paths with different lengths (number of edges) connecting $e_s$ to $e_t$. This plot shows how each of our models performs as the average path length between $e_s$ and $e_t$ increases. Plot (b): as the number of parallel paths between $e_s$ and $e_t$ increases, all of our models' performances decreases. There are many unique number of parallel edges, so we clustered them into $20$ intervals and some of their max values are labelled in the x-axis.}
    \label{fig:testfig}
\end{figure*}
\begin{table}[ht]
\caption{Results for link prediction task in the Freebase+Clueweb dataset, containing 46+1 target relations. The best result from Das et al.~\cite{das2017} was 73.26\%, when considering relations and entity types (the inclusion of labels for unique entities did not improve the MAP). ROP\_ARC3~\cite{yin2018} was able to increase the MAP to 76.16\% by learning entities representations along the path together with the relation sequence. }
\label{tab:results}
\begin{center}
\small{
\begin{tabular}{@{}lc@{}}
\toprule
Model & MAP (\%) \\ \midrule
Single-Model & 70.11 \\
Single-Model + Entity & 71.74 \\
Single-Model + Entity + Types & 72.22 \\
Single-Model + Types & 73.26 \\
ROP\_ARC3 & 76.16 \\
GNN-Relation & 90.77 \\
GNN-Mean & 91.83 \\
GNN-Sum & \textbf{92.32} \\ \bottomrule
\end{tabular}%
}
\end{center}
\end{table}

\section{Analyses and Results}
\label{sec:res}

For each one of the testing instances, our model outputs a ranking over all target relations (in total 46+1 target relations), thus we followed previous works in this dataset and chose the Mean Average Precision (MAP) as our main evaluation metric. However, we restricted it to be MAP@5. The results are presented in Table~\ref{tab:results} and demonstrate how well our models perform under this dataset. Even our simpler model -- \textbf{GNN-Relation} --  was able to achieve $90.77\%$ of MAP over more than 2.1M testing instances. Our best performance, however, came from the \textbf{GNN-Sum} model -- increasing around $16\%$ previous best result (ROP\_ARC3) --  which is somewhat expected, since \emph{Single-Model} also achieved its best results when enhanced with relations and entities types embeddings -- the latter combined via sum.

Despite the outstanding average results, our best model presented some variation in its results when analysed from specific target relations viewpoints. Table~\ref{tab:gnn_sum} shows the top and bottom three target relations, sorted according to average accuracy, a more strict metric than MAP. It is still unclear why there are some relations in which our model has some issues: both TPR and TNR below $75\%$. One possibility is that the negative and positive examples of such troublesome target relations are quite similar in their general labelled graph structure. Nevertheless, even the worst MAP ($83.92\%$) is still higher than the average performance of previous path-based models.

\begin{table}[h]
\centering
\caption{\textbf{GNN-Sum}: Top and bottom three results sorted according to Average Accuracy. Note that the Average Accuracy is closer to TNR due to imbalance between negative and positive testing examples in the dataset. }
\label{tab:gnn_sum}
\begin{adjustbox}{width=0.6\linewidth}
\centering
\begin{tabular}{@{}lcccc@{}}
\toprule
Target relation & MAP & TPR & TNR & Avg. Acc. \\ \midrule
/book/written\_work/original\_language & 98.38 & 93.12 & 97.21 & 96.90 \\
/film/film/directed\_by & 97.92 & 81.00 & 97.31 & 96.13 \\
/music/genre/albums & 97.59 & 87.50 & 95.90 & 95.25 \\ \midrule
/cvg/game\_version/platform & 85.03 & 89.93 & 68.55 & 70.13 \\
/music/artist/origin & 84.32 & 72.46 & 69.06 & 69.31 \\
/people/person/religion & 83.92 & 87.39 & 66.33 & 67.91 \\ \bottomrule
\end{tabular}%
\end{adjustbox}
\end{table}

It is also remarkable that all of our models have no problem dealing with longer paths -- amount of intermediate relations between $e_s$ and $e_t$. In fact, Figure~\ref{fig:testfig}(a) shows that as the Average Path Length increases all of our models tend to perform better. The most significant performance increase is seen on \textbf{GNN-Relation} which starts with very poor performance, due to very little information since relations types embeddings is all that it sees, but quickly stabilises its performance around $92.5\%$ -- after average path length equals to $5$. The other two models also experiment an increase of performance but much less pronounced than the one \textbf{GNN-Relation} experiments. On the other hand, all of our models presented some level of performance decreasing when facing larger numbers of parallel paths between $e_s$ and $e_t$ (see Figure~\ref{fig:testfig}(b)). That is, even though we change the perspective to a graph viewpoint (instead of single paths), our models are much more comfortable dealing with \emph{long} graphs than with \emph{wide} ones.

\section{Conclusions and Future Work}
\label{sec:conc}

The problem of link prediction in a Knowledge Base consists in automatising the discovery of new facts: i.e. learning to predict a missing labelled edge between two nodes. In this context, a \emph{fact} is the connection between two \emph{entities} through a \emph{relation}. This problem arises in scraped structured data, which may miss facts about the existing entities, and also by knowledge bases which were augmented with unstructured data.

Recently, approaches for link prediction have evolved from learning over \emph{edges} to learning over \emph{paths}. In this paper, we go a step further and propose a neural-symbolic based model which learns directly over \emph{graphs}, enabling reasoning over many paths at once. We achieve this by feeding the minimal subgraph containing all known paths between two entities to a GNN, a deep learning architecture which learns real-valued vector representations for nodes (entities) and edges (facts) in a graph. After the GNN learns a representation for the missing fact, we use a MLP to decode it, enabling us to predict the relation it represents.

Our results surpassed all previous path-based models in a dataset from Freebase+ClueWeb containing over 3M entities -- in the best scenario we increased the MAP metric by $16\%$ over $46+1$ target relations. However, there are two main points to be addressed in the future: (i) our model relies on matrix multiplications to produce embeddings' updates, and can be improved to deal with huge graphs; (ii) we currently update entities types and relations embeddings at the end of the pipeline, with the Softmax Cross-Entropy loss; Yin et al.~\cite{yin2018} were able to improve previous models' results by updating entities representations along the entire path; in this sense our model could be adapted to perform updates in the same manner. 

%
%
%
\bibliographystyle{splncs04}
\bibliography{samplepaper}
\end{document}